\documentclass[conference]{IEEEtran}

\usepackage[numbers,sort&compress]{natbib}
\usepackage{multicol}
\usepackage[bookmarks=true,pagebackref]{hyperref}
\usepackage{xspace}

\newcommand{\midnet}{waypoint network\xspace}
\newcommand{\midneta}{WayNet\xspace}
\newcommand{\latentmap}{memory proxy map\xspace}
\newcommand{\latentmapa}{MPM\xspace}

\usepackage{comment}
\usepackage{tabularx}
\usepackage{booktabs}
\usepackage{amsmath}
\usepackage{graphicx}
\usepackage{subfigure}
\let\labelindent\relax
\usepackage{enumitem}
\usepackage{wrapfig}
\usepackage{amssymb}
\usepackage{pifont}
\usepackage[ruled,vlined]{algorithm2e}
\usepackage{mathtools}

\usepackage{amsmath}
\usepackage{soul}
\usepackage{multirow}
\usepackage[noend]{algpseudocode}
\usepackage{subfigure}
\usepackage{fancyhdr}

\title{Feudal Networks for Visual Navigation}
\author{\authorblockN{\large Faith Johnson}
\authorblockA{\small Rutgers University\\
faith.johnson@rutgers.edu}
\and
\authorblockN{\large Bryan Bo Cao}
\authorblockA{\small Stony Brook University\\
boccao@cs.stonybrook.edu}
\and
\authorblockN{\large Ashwin Ashok}
\authorblockA{\small Georgia State University\\
aashok@gsu.edu}
\and
\authorblockN{\large Shubham Jain}
\authorblockA{\small Stony Brook University\\
jain@cs.stonybrook.edu}
\and
\authorblockN{\large Kristin Dana}
\authorblockA{\small Rutgers University\\
kristin.dana@rutgers.edu}
}

\newcommand{\cmark}{\ding{51}}%
\newcommand{\xmark}{\ding{55}}%

\begin{document}

\maketitle

\begin{abstract}
Visual navigation follows the intuition that humans can navigate without detailed maps. 
A common approach is interactive exploration while building a topological graph with images at nodes that can be used for planning.  Recent variations learn from passive videos and can navigate using complex social and semantic cues. However, a significant number of training videos are needed, large graphs are utilized, and scenes are not unseen since odometry is utilized.  
We introduce a new  approach to visual navigation using feudal learning, which employs a hierarchical structure consisting of a worker agent, a mid-level manager, and a high-level manager. Key to the feudal learning paradigm, agents at each level see a different aspect of the task and operate at different spatial and temporal scales.  
Two unique modules are developed in this framework. For the high-level manager, we learn a {\it \latentmap} in a self supervised manner to record prior observations in a learned latent space and avoid the use of graphs and odometry. For the  mid-level manager, we develop a {\it \midnet} that outputs intermediate subgoals imitating human waypoint selection during local navigation. This \midnet is pre-trained using a new, small set of teleoperation videos that we make publicly available, with training environments  different from testing environments.  The resulting feudal navigation network achieves near SOTA performance, while providing a novel no-RL, no-graph, no-odometry, no-metric map approach to the image goal navigation task.

\end{abstract}

\section{Introduction}

Visual navigation is motivated by the idea that humans likely navigate without ever building detailed 3D maps of their environment. 
In psychology, the concept of cognitive maps and graphs \cite{tolman1948cognitive, chrastil2014cognitive,peer2021structuring, epstein2017cognitive} formalizes this intuition, and experiments have shown the potential validity of the idea that humans build approximate graphs of their environment encoding relative distances between landmarks.  In vision and robotics, these ideas have translated to the construction of topological graphs and maps 
based solely on visual observations.
Visual navigation paradigms seek new representations of environments that are rich with semantic information, easy to dynamically update, and can be constructed faster and more compactly than full 3D metric maps \cite{savinov2018semi,chaplot2020neural,mirowski2018learning,Savarese-RSS-19,gervet2023navigating}.

Reinforcement Learning (RL) is often used for navigation in known environments and has clear success in training games where the reward is well-defined \cite{mnih2015human, mnih2013playing, lillicrap2015continuous}. Distance-to-goal has become a reward proxy in many recent works that use an RL framework for navigation \cite{shah2021ving, eysenbach2019search}. We take inspiration from Feudal RL \cite{vezhnevets2017feudal} to craft our visual navigation framework.
Feudal RL decomposes a task into sub components and can provide significant advantages for both training and performance 
that we find particularly well-suited for the task of visual navigation. 
The feudal framework identifies {\it workers} and  {\it managers}, and allows for multiple levels of hierarchy (ie. mid-level and high-level managers). Each of these entities observes different aspects of the task {\it and} operates at a different temporal or spatial scale.  For navigation in unseen environments, this dichotomy is ideal  to make the overall task more manageable. 
The worker-agent can focus on local motion, but manager-agents can direct 
navigation and assess when it is time to move to new regions.

However, in this work we focus on visual navigation with no odometry and in unknown environments, so distance-based reward is not readily available to train an RL agent. 
Recent work called NRNS \cite{hahn2021no}, which builds a topological graph using both images and odometry information, has challenged the need for RL and instead uses a distance estimator by training with passive videos. 
Inspired by NRNS, we also use ``no-RL'' so that no policy is learned via reinforcement/interaction with an environment. We take advantage of the feudal RL network structure and show its benefits under supervised and self-supervised no-RL learning paradigms. 
Our approach uses no metric maps, no graphs, and no odometry information,  resulting in a lightweight easy-to-train framework that still has high/SOTA performance in image-goal navigation tasks.

\begin{figure}
    \centering
    \includegraphics[width=3.6in]{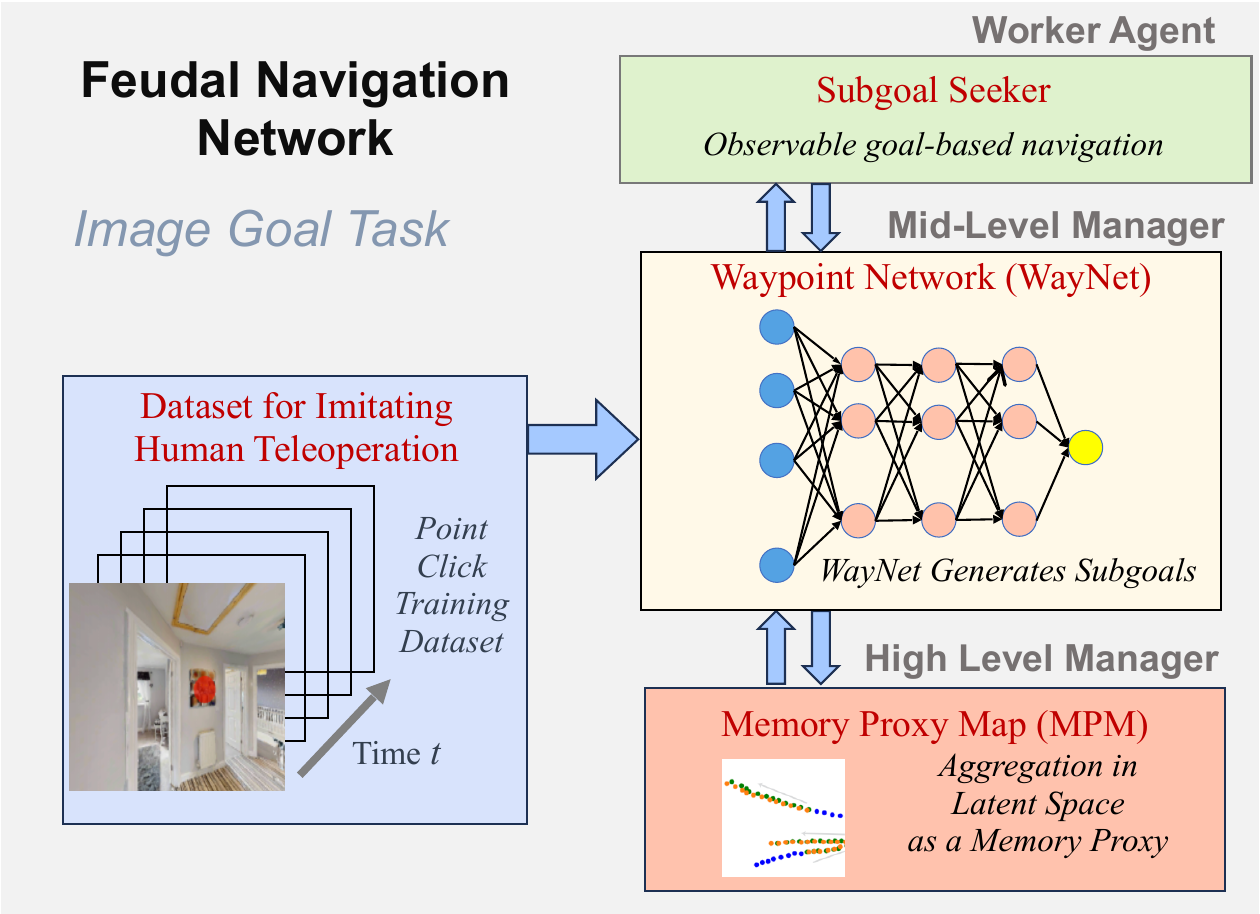}
   \vspace{-12pt}
    \caption{Feudal Navigation Network (FeudalNav), providing a no-RL, no-odometry, no-graph, and no-metric map visual navigation agent for the image-goal task on previously unseen environments. The three main components are: (1) a high level manager that creates a \latentmap (\latentmapa)  to use as an aggregate observation to make high-level navigation decisions, (2) a mid-level manager \midnet (\midneta) that mimics human teleoperation by predicting visible points in the environment to guide worker agent exploration, and (3) a low level worker that finds the subgoal using robust point matching.}
    \label{fig:overview}
    \vspace{-10pt}
\end{figure}

Key to our approach is representing the traversed environment with a learned latent map (instead of a graph) that acts as a memory proxy during navigation.  This {\it \latentmap (\latentmapa)} is obtained using self-supervised training.
The high-level manager maintains the \latentmapa as the agent navigates in novel environments and polls its density to determine when a region is well-explored and movement away from the current region is desired. A second key aspect to our approach is a {\it \midnet (\midneta)} for the middle-level manager which outputs visible sub-goals for the worker agent to move towards.  We train \midneta to imitate human teleoperation and collect a dataset with a human teleoperator tasked with  point click navigation through a set of Gibson environments \cite{xiazamirhe2018gibsonenv} in Habitat AI \cite{habitat19iccv}  (a simulation environment comprised of scans of real scenes). The  resulting ``image and point-click''  pairs are used to train \midneta in a set of environments different from the image-goal testing environments. The intuition is that when humans navigate a simulated environment using point-click teleoperation, they use a skill of choosing a single point in the observation to move toward. For example, the chosen point may be toward the end of a hallway, toward a door, or further into a room.  We demonstrate that this skill is easily learnable and generalizes to new environments with zero-shot transfer. Our dataset of images and point-clicks is made publicly available with this paper.

%



Our contributions are fivefold:
\begin{enumerate}
    \item Hierarchical navigation framework (FeudalNav) using agents operating at different spatial and temporal scales
    \item A lean, no-graph navigation approach using a self-supervised \latentmap (\latentmapa)
    \item \midneta, a  \midnet for learned local navigation
    \item A dataset \footnote{Dataset can be accessed here: \\ \href{https://huggingface.co/datasets/visnavdataset/lavn}{https://huggingface.co/datasets/visnavdataset/lavn}.} of 103K point-clicks from human teleoperation in multiple environments made publicly available and useful for training a \midnet  
    \item SOTA performance on the image-goal task in Habitat indoor environments (testing and training on distinct environments)
\end{enumerate}


\begin{figure*}
  \centering
    {\includegraphics[width=\textwidth]{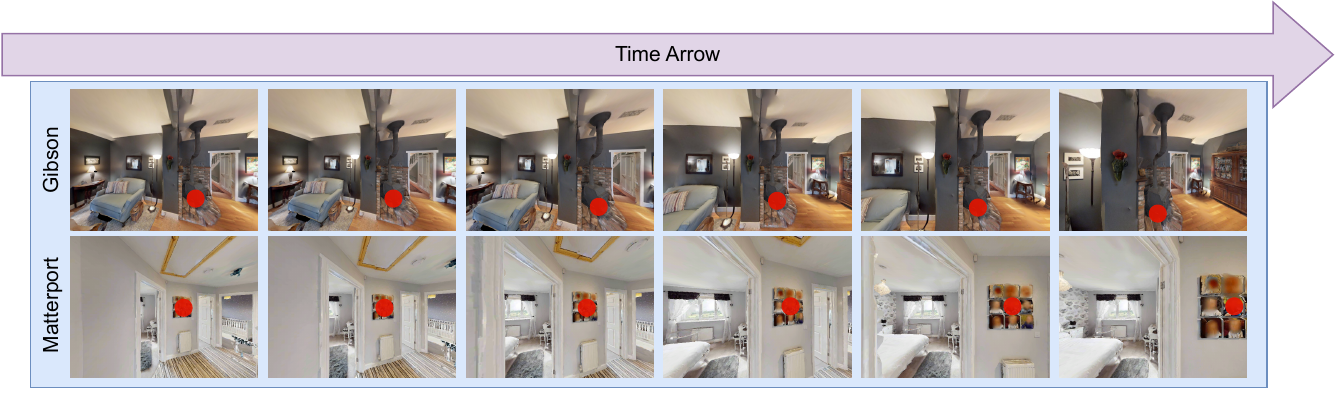}}
    \vspace{-15pt}
  \caption{Samples of RGB images in our human navigation dataset captured from an ego-centric camera in both Gibson \cite{xiazamirhe2018gibsonenv} and Matterport \cite{Matterport3D} environments. Video frames start from left to right. A human provides point-click guidance for robot visual navigation visualized with red dots.}
  \label{fig:vis}
  \vspace{-10pt}
\end{figure*}
%

\section{Related Work}

\subsection{Visual Navigation}




Visual navigation aims to build representations that incorporate the rich information of scene images, interjecting image-based learning in traditional mapping/planning  navigation frameworks \cite{Gupta_2017_CVPR, chaplot2019learning, devo2020towards, shah2021ving,seymour2021maast}.
Visual navigation methods can be categorized based on how environments are represented. Early work focused on creating full metric maps of a space using SLAM augmented by images \cite{chaplot2019learning,chaplot2020neural}. 
While full metric maps can be ideal, especially if the space can be mapped before planning, the representations are computationally complex.
Topological graphs and maps can lighten this load  and provide image data at nodes and relative distances at edges \cite{savinov2018semi,chen2019behavioral}.
While easier to build, these methods have the potential for large memory requirements, especially if new nodes are added every time an agent takes an action in the environment \cite{shah2021ving,shah2022offline,he2023metric}.

One solution to this problem is sparser topological graphs \cite{hahn2021no,shah2021rapid} that retain information useful for navigation.
Visual features of unexplored next-nodes are sometimes predicted or hallucinated  \cite{hahn2021no,he2023metric}.
Some methods go a step further by pairing semantic labels with the graph representation \cite{kim2023topological}. However, these methods break down in environments that are sparsely furnished, have many duplicate objects, or contain uncommon objects that may not appear in popular object detection datasets. Another solution is to only use graphs during training to build 2D embedding space representations of environments, i.e.\ potential fields or functions, that preserve important physical \cite{morin2023one, ramakrishnan2022poni}, visual \cite{henriques2018mapnet}, or semantic \cite{georgakis2021learning,chaplot2020object,al2022zero} relationships between regions in the environment.  Our work more closely aligns with this line of work,
but we do not use any graph networks or inference and instead
 build our 2D latent map, the \latentmap, using self-supervision. 
Additionally, unlike many of the methods above,   we do not require the agent to have information about the test environment before deployment, use reinforcement learning, or learn metric maps.






\subsection{Feudal Learning}
Feudal learning originated as a reinforcement learning (RL) framework \cite{dayan1992feudal,vezhnevets2017feudal}. Researchers have explored RL for the image-goal visual navigation task \cite{zhu2017target}, most notably using external memory buffers \cite{kumar2018visual, fang2019scene, beeching2020egomap, mezghan2022memory}.
However, it still suffers from several issues such as sample inefficiency, handling sparse rewards, and the long horizon problem \cite{fujimoto2021minimalist,le2018hierarchical}. 
Feudal reinforcement learning, characterized by it's composition of multiple, sequentially stacked agents working in parallel, arose to combat these issues using temporal or spatial abstraction, mostly in simulated environments \cite{vezhnevets2017feudal} where its effects can be more easily compared to other methods. Some of these task hierarchies are hand defined \cite{vezhnevets2020options}, while others are discovered dynamically with \cite{chen2020ask} or without \cite{li2020hrl4in} human input. 

The feudal network paradigm has been adopted by other learning schemas outside of RL in recent years. In navigation, hierarchical networks are commonly used to propose waypoints as subgoals during navigation \cite{chane2021goal}, typically working in a top-down view of the environment \cite{xu2021hierarchical} with only two levels of agents \cite{wohlke2021hierarchies}. Our work uses multiple agent levels, operates in the first person point of view for predicting waypoints, and reaps the benefits of this feudal relationship without using reinforcement learning. 

\begin{figure}[t]
  \centering
  \begin{minipage}{1\linewidth}
  \subfigure[Frames]{\includegraphics[width=0.49\textwidth]{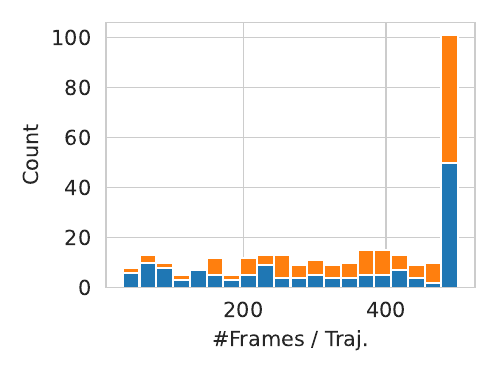}}\label{fig:f}
  \subfigure[LM]{\includegraphics[width=0.49\textwidth]{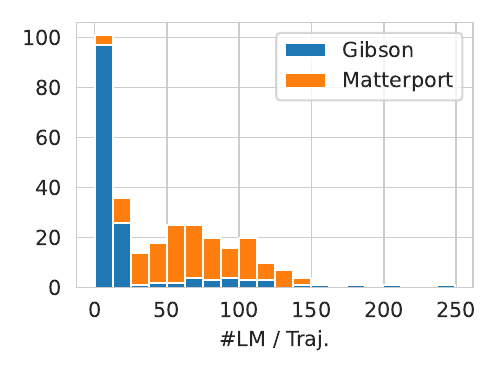}}\label{fig:lm}
  \caption{Stacked histograms depicting $\#$Frames and $\#$LM per trajectory. (a) the majority of trajectories consists of the maximum $\#$Frames 500. The number of human point-click waypoints is equal to the number of frames for each scene. (b) most trajectories in the Gibson rooms consists of only a small number of annotated 
  landmarks (e.g. $<25$), while it increases in Matterport due to its larger and more complicated environments. 
  (LM denotes landmark.) }
  \label{fig:datahist}
  \end{minipage}
\end{figure}

\begin{figure*}
    \centering
    \includegraphics[width=\linewidth]{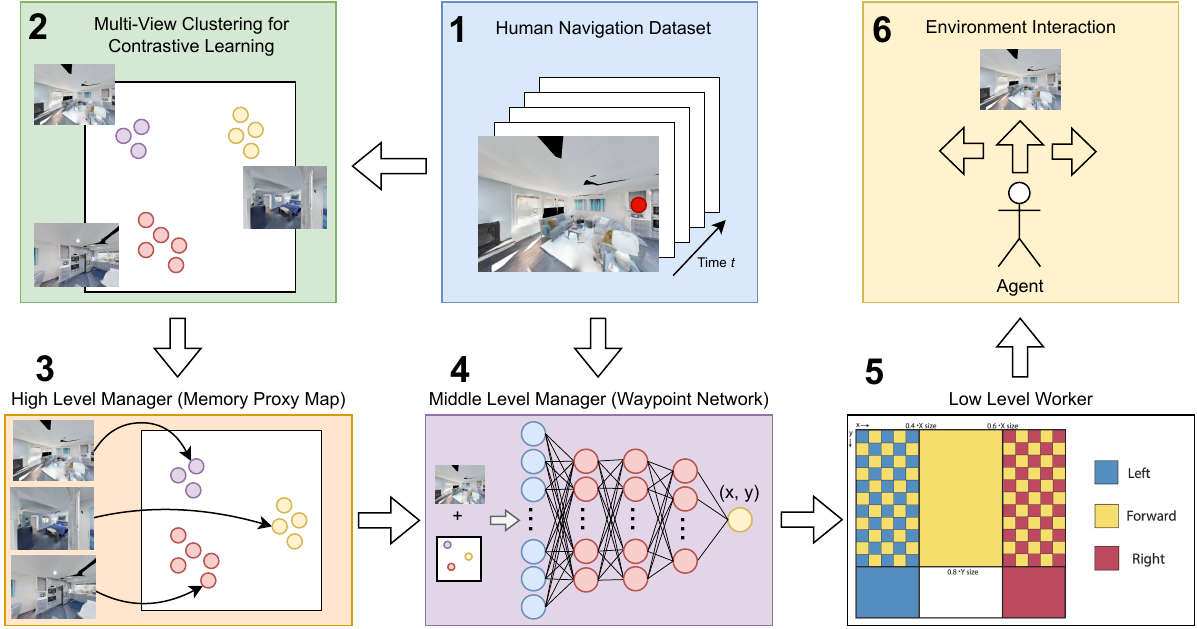}
    \vspace{-12pt}
   \caption{\textbf{1:}  Point click data is collected while human teleoperators direct agent exploration of different environments. The resulting set of point-image pairs comprise the human navigation dataset. \textbf{2:} From this dataset, we find clusters of observations based on feature similarity. \textbf{3:} These clusters are used to provide positive pairs to train the navigation memory module that serves as the high level manager (HLM) for our navigation agent. During test time, this HLM creates a map of historical agent locations (\latentmap) in the learned space. \textbf{4:} These maps are created for the human navigation dataset and used to train the mid-level manager. During training, the \latentmap and the current observations are used to predict human-like point click supervision to guide environment exploration. \textbf{5:} Based on this point click guidance, the worker executes low level actions directly in the simulated environment. (See Figure \ref{fig:workerActions}) \textbf{6:} During test time, these low level actions guide agent movement and produce new observations as input for the upper levels of the hierarchy.  }
    \label{fig:processdiagram}
    \vspace{-10pt}
\end{figure*}

\section{Human Navigation Dataset}

Humans are already adept navigators, but teaching agents to learn proficient navigation policies often requires distilling complex concepts and strategies into mathematical equations to be used for reward or loss functions. However, it is possible to avoid this lengthy process by implicitly learning policies directly from observations of human behavior. We collect videos of human driven trajectories through 
multiple simulated and real world environments to be used for this purpose. 


\subsection{Environments}
We provide human point-click waypoint and landmark annotations during navigation for a selection of the Gibson \cite{xiazamirhe2018gibsonenv} and Matterport \cite{Matterport3D} environments in Habitat-Sim \cite{szot2021habitat,habitat19iccv} to facilitate virtual environment exploration, as shown in Figure \ref{fig:vis}. 
Habitat is already used to train many visual navigation methods, so having in-domain, landmark-aware data will allow for direct extensions of these methods for greater ease of model comparison. Additionally, collecting data in simulation allows for greater collection volume and lower data processing time. 




\begin{table}
  \begin{center}
\begin{tabular}{|c|cccc|}
\hline
 & $\#$Traj. & $\#$LM & Tot. $\#$Frames & $\#$Frames / Traj.\\
\hline
\hline

G & 150 & 3,744 & 48,130 & 321 \\
MP & 150 & 10,537 & 55,368 & 369 \\
\hline
All & 300 & 14,281 & 103,498 & 345 \\
\hline

\end{tabular}
\end{center}
\caption{Statistics of our Human Navigation Dataset. We detail the number of trajectories, landmark annotations, total number of frames and average number of frames per trajectory. Each trajectory is recorded in a distinct environment. LM: landmarks, G: Gibson, MP: Matterport.}
\label{tab:stats}
\vspace{-25pt}
\end{table}


\subsection{Data Collection and Representation}
Instead of having agents reason over low level motor actions, we leverage the simulator to enable reasoning over high level actions, like ``turn left", ``turn right", and ``go forward". These simplified actions also allow for a simplification of the human feedback needed to collect this trajectory data.
To start, the human navigator is placed at a random, navigable point in an environment, and tasked with visiting as many rooms as possible while identifying distinct waypoints in each room. The current first person observation of the environment is shown on a screen, and the human moves through the environment by left clicking on this image in the direction they would like to move. There are three discrete action choices for moving around in the environment: moving forward $0.25$m, turning left $15^{\circ}$, and turning right $15^{\circ}$. We map each human point click to a combination of these three actions as specified in Figure \ref{fig:workerActions}, execute them in the environment, then give the human the resulting new frame to continue the navigation process. To specify a landmark (or useful point for localization), the navigator right clicks on the observation image over the object or region of space they choose. 

We record the sequence of observations (RGBD images), actions, navigation point clicks, and environment locations (X,Y,Z coordinates) as the navigator moves through each space. 
A list of pairs of  point-click coordinates and their corresponding image observations is also provided so that landmarks may be easily identified in the scenes. Each human-guided trajectory is terminated after 500 actions or when the annotator has navigated through the entirety of the space, whichever comes first. We summarize the data distribution in Figure \ref{fig:datahist} and dataset statistics in Table \ref{tab:stats}.

\section{Methods}
We start with an overview of our feudal navigation agent (FeudalNav), which is split into three levels, in Figure \ref{fig:processdiagram}. First, we collect human navigation point-click supervision trajectories from several simulated environments as detailed in Section III. Then, we find clusters in that data corresponding to each unique visual area. The high level manager network learns a latent space in a self-supervised manner that serves as an approximate distance preserving memory module (\latentmap) for navigation from this clustered data. The mid-level manager network (\midnet) uses this memory representation and the current observation to mimic human navigation policies by predicting a point in the environment to move towards. 
The low level worker agent uses this predicted point to choose which actions to execute in the environment. We also add several goal-directed modules to our architecture that utilize Superglue \cite{sarlin2020superglue} in order to test FeudalNav on the image-goal navigation task.

\subsection{High-Level Manager: Memory}
We contrastively learn a latent space that preserves an approximate distance between images to be used to build an aggregate \latentmap (\latentmapa). We learn this self-supervised latent space using a modified implementation of SMoG \cite{pang2022smog} that combines instance level contrastive learning and clustering methods. We choose this model because its momentum grouping gives it the capability of conducting both instance and group level contrastive learning simultaneously, which allows it to avoid the pitfall of other contrastive methods of using false negative pairs. We add further modifications to model training in order to conduct navigation-aware, self-supervised contrastive learning.

Instead of using typical augmentation methods during training, we rely on the slight variations introduced through multiple real world views to learn robust image representations. To determine positive pairs, we dynamically build clusters of images in the environment based on visual similarity. To do this, we choose representative images to serve as cluster centers for each room or distinct environment area during navigation. As an agent traverses through an environment, the current observation is compared to the memory bank of previously seen cluster centers  using Superglue \cite{sarlin2020superglue}, a feature matching method for robust keypoint matching.  If the feature matching confidence provided by Superglue is above a threshold $\psi$, it is added to the corresponding cluster with the highest confidence score.  If it is below that threshold, it is added as a new cluster center. We build these clusters per environment for all training trajectories and randomly sample positive pairs from each cluster to train the network. 

During inference, we sequentially place observation images in this latent space to dynamically build a \latentmap of previously visited locations. We find clusters in this latent space in a similar way as during training, but using the contrastively learned image features instead of utilizing the Superglue network. 
If the mean squared error (MSE) between the learned features of any of the existing cluster centers $C_{f}$ and the features $I^t_{f}$  from the new observation at time $t$ is below a threshold $\alpha$, it is considered to be part of the cluster with the lowest MSE. That is, if $\mathit{MSE}(C_{f},I^t_{f}) < \alpha$ it is added to the cluster with the lowest MSE. However, if $\mathit{MSE}(C_{f},I^t_{f}) > \alpha$, it is added as a new cluster center.


To update the \latentmapa, we use the isomap algorithm to project each discovered cluster center from the image feature space to a 2-D latent space that preserves the distance between the learned features. 
Then, a region around each projected cluster center in the 2D map is incremented and the radius of the region indicates the number of historical observations assigned to each cluster. 
This process yields a map with circular regions of varying sizes corresponding to how much exploration has occurred around each location.
By representing the map with three channels, analogous to RGB, the increments are made with 3-channel values designating a unique ``color'' to each cluster center in order to help localize the agent.
The more a room in an environment is explored, the larger the 
region in the \latentmapa corresponding to this area. Looking at this coded cluster density gives the network an idea of how much exploration is left to do in a given space and  signals when it is time to move to a new area 
using 
the relative image feature distance between the current observation and representative samples of previously seen areas, 
without the need for metric maps or SLAM. 
The approach is an approximate memory proxy, that is demonstrably effective and efficient for the image-goal navigation task. 


\subsection{Mid-Level Manager: Direction}
For this module, we leverage human knowledge to learn optimal navigation policies directly from demonstrations collected in our human navigation dataset without the use of RL. Using the current RGBD observation and the \latentmap from the high level manager as input, we finetune Resnet-18 \cite{he2016deep} to predict the pixel coordinate directing the navigation agent's motion in the environment. We alter the first layer of Resnet so that it can accept a seven channel input and concatenate the RGBD observation and the (RGB) \latentmap together along the channel dimension to use as input. We also create a second version of the mid-level manager that takes in multiple of these RGBD-Map compound inputs in order to give the network more context about recently visited areas. For this network version, which we use in Stacked FeudalNav, we use the RGBD-Map input of the previous two frames and the current observation concatenated together along the channel dimension as input to \midneta to predict the next navigation waypoint.

The key intuition here is that the high level manager reasons about the navigation task at a  more global view and 
coarser grained spatial scale than the mid-level manager, that has a first person, more fine-grained view of the environment. Breaking the problem into these two spatial scales introduces spatial abstraction which allows these two networks to work in tandem to solve smaller pieces of the overall problem. Stacked FeudalNav's input that covers three time steps further introduces a level of temporal abstraction between the high level manager, which now operates at a coarser grained time scale, and the mid level manager, which has a more fine grained time scale, which further breaks down the original navigation task. These simplifications allow for faster training with smaller amounts of data.

To enable goal-directed navigation, we again utilize the Superglue \cite{sarlin2020superglue} network for keypoint matching. 
For each new observation, the network predicts a waypoint for exploration. Concurrently, Superglue computes keypoint matches between the current observation and a goal image. If the confidence of this keypoint match is above a threshold $\beta$, the average of the matched keypoints is used in the navigation pipeline instead of the waypoint prediction. If this is not the case, then the originally predicted exploration waypoint is used instead.

\subsection{Low-Level Worker: Action}
\begin{figure}
    \centering
    \includegraphics[width=0.85\linewidth]{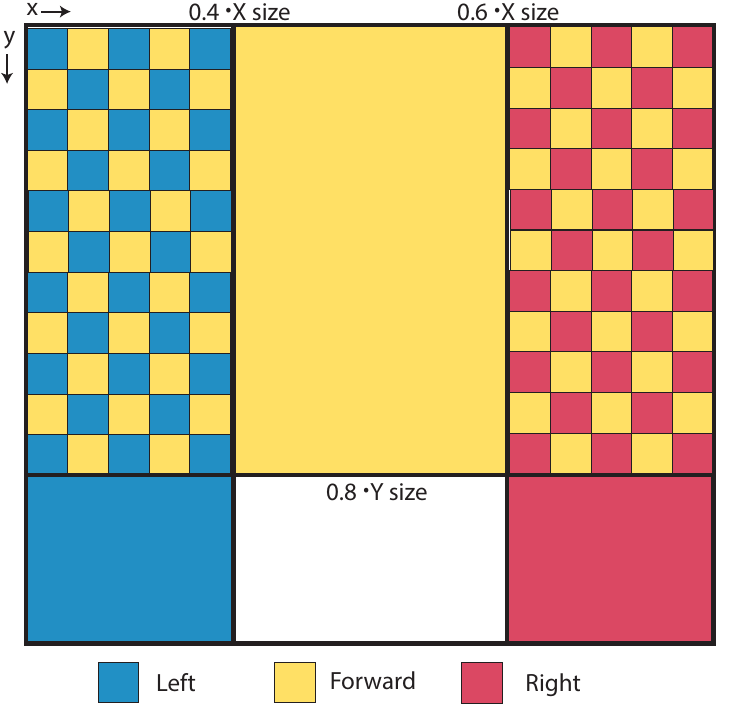}
    \vspace{-5pt}
    \caption{Map showing mid-level manager point click locations map to which simulator actions for the low level worker. The bottom left grid (blue) and the bottom right grid (red) correspond to  ``turn left" and 11turn right" respectively. The center middle grid (yellow) corresponds to ``move forward". The top left grid (blue and yellow) and the top right grid (red and yellow) correspond to the joint action of turning (left for blue and right for red) and moving forward sequentially. The two vertical delineations are at 0.4 and 0.6 times the x dimension of the observation respectively. The horizontal line is drawn at 0.8 time the y dimension of the observation.}
    \label{fig:workerActions}
    \vspace{-10pt}
\end{figure}

The low level worker agent takes actions in the environment based on the waypoint  predicted by the mid-level manager. We define three environment level actions: ``turn left by $15$ degrees", ``turn right by $15$ degrees", and ``move forward by 0.25 meters (m)". The observation is split into a grid with three columns and two rows, shown in Figure \ref{fig:workerActions}, and the location of the predicted point determines which actions are taken. The bottom left (blue) and bottom right (red) grids correspond to ``turn left" and ``turn right" respectively. The top middle section (yellow) corresponds to ``move forward". The top left (blue and yellow) and top right (red and yellow) grids correspond to the action sequence of first turning (left or right respectively) and then moving forward. Predicting a waypoint in the bottom middle white box results in no action being taken. The agent chooses to stop when the Superglue \cite{sarlin2020superglue} confidence threshold for matching goal image features to the current observation is above the threshold $\beta$ and the depth measurement from the RGBD observation indicates that the middle of the matched keypoints has a depth less than or equal to $1$m.

We introduce temporal abstraction into the relationship between the mid-level manager and the low level worker by enforcing that the two networks will operate at different temporal scales. The worker makes $t$ actions in the environment for each exploration waypoint provided by the mid-level manager. For the first of these $t$ actions, it directly uses the predicted waypoint as its supervisory signal. We take a crop around the waypoint in the original observation and use Superglue \cite{sarlin2020superglue} to match it to a location in the new observations for the remaining $t-1$ actions. The average of the keypoint matches is used as the waypoint to determine the new actions for each successive time step. We also add a level of stochasticity to the worker agent. With probability $\epsilon$, it takes random actions in the environment that do not necessarily correspond to the waypoint supervision from the mid-level manager in order to further promote exploration.

\section{Experiments}

\subsection{Image-Goal Navigation Task}
We test the performance of FeudalNav on the image-goal task. To start, the agent is placed in a previously unseen environment and given an RGBD image observation of the first person view of their surroundings. It is also given a goal image (RGB) of an object or region in the scene to find. All images are $480 \times 640$ pixels with $120^\circ$ field of view. A trial terminates if the agent is able to get within $1$m of the location of the goal image or the agent takes $500$ actions in the environment.
We define three environment level actions: ``turn left $15^\circ$", ``turn right $15^\circ$", and ``move forward $0.25$m" for the agent to choose from. The combined action choice of turn+forward represented by the upper left and right quadrants in Figure \ref{fig:workerActions} counts as two actions towards the agent action limit.
Each agent trajectory is evaluated on success rate, which measures whether or not the agent has reached the goal, and SPL, which is a measure of success weighted by inverse path length. 
\[ SPL = \frac{1}{N}\sum^N_{i=0}S_i\frac{l_i}{\max(l_i,p_i)}\]
where $N$ is the total number of trajectories considered, $S_i$ is an indicator variable for success, $l_i$ is the optimal (shortest) geodesic path length between the starting location and the goal, and $p_i$ is the actual path length the agent traveled.

\subsection{Training Procedure}


We train our networks using the human navigation dataset detailed in Section III.
We use a subset of 117 trajectories with a total of 36834 frames. 
For the Superglue \cite{sarlin2020superglue} confidence threshold when creating the clusters from the human navigation dataset or matching goal image features to the current observation, we choose $\psi = \beta = 0.7$. We choose $\alpha = 300$ for the threshold for assigning a learned image feature from SMoG \cite{pang2022smog} to an existing cluster center of the \latentmap. The mid-level manager and the low level worker operate at different time scales, so the worker takes $t=2$ steps in the environment for every one manager waypoint. Additionally, the worker takes random actions in the environment with a probability of $\epsilon = 0.1$.

\begin{figure}
    \centering
    \includegraphics[width=\linewidth]{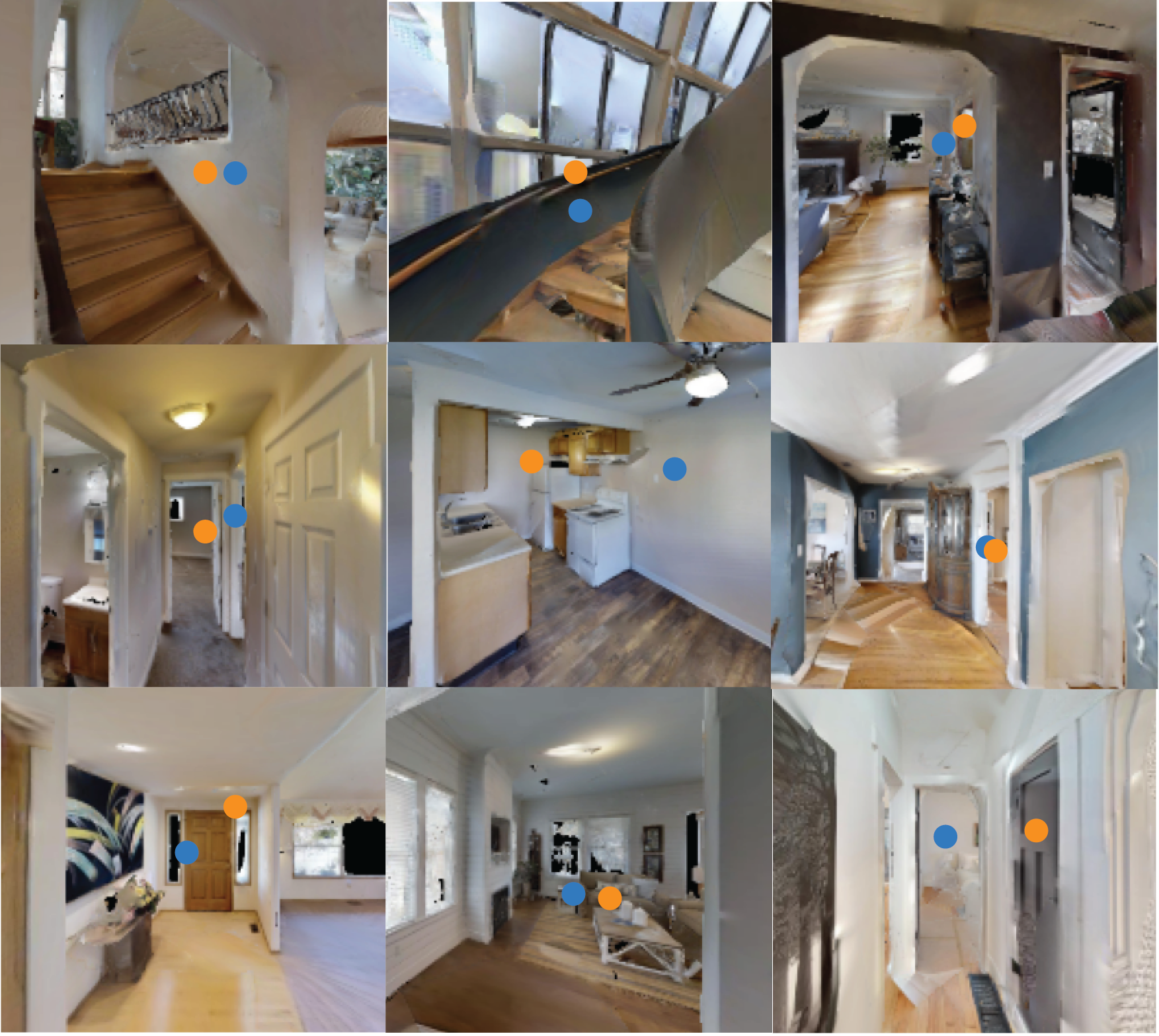}
    \vspace{-15pt}
    \caption{(Best viewed zoomed) We show qualitative results for the waypoints predicted by \midneta (blue) juxtaposed with the ground truth human click points from the human navigation dataset (orange). Note that the majority of the samples show high overlap between the two. When they diverge, the \midneta waypoints still lead to navigably feasible areas in each observation.}
    \vspace{-10pt}
    \label{fig:mlmPoints}
\end{figure}

\subsection{Testing Procedures}
We test FeudalNav using the testing procedure and baselines outlined in NRNS \cite{hahn2021no}. 
Testing trajectories come from a publicly available set of Gibson \cite{xiazamirhe2018gibsonenv} environments listed in \cite{hahn2021no}. They consist of approximately $6k$ point pairs (start and goal locations) that are uniformly sampled from fourteen environments and divided equally across the following categories. Straight trajectories involve two points where the ratio between the shortest geodesic distance and the euclidean distance between the points is less than 1.2 and the difference between the ground truth start and stop orientations is less than $45^\circ$. Curved trajectories involve point pairs where either of these conditions cannot be satisfied. Easy trajectories involve points that are $[1.5-3m)$ apart, medium trajectories have endpoints that are $[3-5m)$ apart, and hard trajectories have endpoints that are $[5-10m)$ apart by euclidean distance. 

We compare our method performance using a flat RL DD-PPO \cite{wijmans2019dd} trained for varying lengths of time directly in the simulator, behavior cloning (BC) with a resnet-18 backbone and either a GRU or a metric map from \cite{hahn2021no}, as well as with NRNS itself \cite{hahn2021no}. We believe this is the best baseline for our method due to the fact that they train from passive videos, use a (relatively) low amount of training data compared to other SOTA, minimally use graphs, and do not train in simulation. 
There are other recent works that test on the image-goal task that we believe are an unfair comparison with our work due to their use of methods (scene reconstruction \cite{kwon2023renderable}, transfer learning \cite{al2022zero}, long training times ie. 53 GPU days \cite{yadav2022offline}, panoramic image input \cite{kim2023topological}) that are outside the scope of this paper due to their computationally heavy nature and/or requiring testing in previously seen environments. 




\begin{table*}[]
    \centering
    \begin{tabular}{|c|p{14em}|c|c|c|c|c|c||c|c|}
    \hline
    \multirow{2}{2em}{Path Type} & \multirow{2}{7em}{Model} & \multicolumn{2}{c|}{Easy} & \multicolumn{2}{c|}{Medium} & \multicolumn{2}{c||}{Hard} & \multicolumn{2}{c|}{Average}\\
     & & Succ$\uparrow$ & SPL$\uparrow$ & Succ$\uparrow$ & SPL$\uparrow$ & Succ$\uparrow$ & SPL$\uparrow$ & Succ$\uparrow$ & SPL$\uparrow$\\
    \hline
    \hline
    \multirow{9}{3em}{Straight} & RL (10M steps) * \cite{wijmans2019dd} &  10.50 & 6.70 & 18.10 & 16.17 & 11.79 & 10.85 & 13.46 & 11.24 \\
    & RL (extra data + 50M steps) * \cite{wijmans2019dd}  & 36.30 & 34.93 & 35.70 & 33.98 & 5.94 & 6.33 & 25.98 & 25.08 \\
    & RL (extra data+100M steps) * \cite{wijmans2019dd} & 43.20 & 38.54 & 36.40 & 34.89 & 7.44 & 7.20 & 29.01 & 26.88 \\
    & BC w/ ResNet + Metric Map \cite{hahn2021no}   &  24.80 & 23.94 & 11.50 & 11.28 & 1.36 & 1.26 & 12.55 & 12.16 \\
    & BC w/ ResNet + GRU \cite{hahn2021no} &  34.90 & 33.43 & 17.60 & 17.05 & 6.08 & 5.93 & 19.53 & 18.80 \\
    & NRNS w/ noise \cite{hahn2021no}  & 64.10 & 55.43 & 47.90 & 39.54 & 25.19 & 18.09 & 45.73 & 37.69 \\
    & NRNS w/out noise \cite{hahn2021no} & \textbf{68.00} & \textbf{61.62} & 49.10 & \textbf{44.56} & 23.82 & 18.28 & 46.97 & \textbf{41.49} \\
    & FeudalNav (Ours) & 61.40 & 51.81 & 50.30 & 40.10 & 31.02 & \textbf{23.33} & 47.57 & 38.41  \\
    & Stacked FeudalNav (Ours) & 65.90 & 48.50 & \textbf{51.00} & 29.22 & \textbf{33.62} & 14.49 & \textbf{50.17} & 30.74  \\
    \hline
    \multirow{9}{3em}{Curved} & RL (10M steps) * \cite{wijmans2019dd} &7.90 & 3.27 & 9.50 & 7.11 & 5.50 & 4.72 & 7.63 & 5.03 \\
    & RL (extra data + 50M steps)* \cite{wijmans2019dd} & 18.10 & 15.42 & 16.30 & 14.46 & 2.60 & 2.23 & 12.33 & 10.70 \\
    & RL (extra data+100M steps)* \cite{wijmans2019dd} & 22.20 & 16.51 & 20.70 & 18.52 & 4.20 & 3.71 & 15.70 & 12.91 \\
    & BC w/ ResNet + Metric Map \cite{hahn2021no} &3.10 & 2.53 & 0.80 & 0.71 & 0.20 & 0.16 & 1.37 & 1.13 \\
    & BC w/ ResNet + GRU \cite{hahn2021no} &3.60 & 2.86 & 1.10 & 0.91 & 0.50 & 0.36 & 1.73 & 1.38 \\
    & NRNS w/ noise \cite{hahn2021no}  &27.30 & 10.55 & 23.10 & 10.35 & 10.50 & 5.61 & 20.30 & 8.84 \\
    & NRNS w/out noise \cite{hahn2021no} &35.50 & 18.38 & 23.90 & 12.08 & 12.50 & 6.84 & 23.97 & 12.43 \\
    & FeudalNav (Ours) & 41.30 & 19.51 & 32.60 & 17.10 & 18.60 & \textbf{10.88} & 30.83 & \textbf{15.83} \\
    & Stacked FeudalNav (Ours) & \textbf{56.40} & \textbf{21.37} & \textbf{44.10} & \textbf{17.41} & \textbf{21.00} & 7.30 & \textbf{40.50} & 15.36 \\
    
    \hline
    \end{tabular}
    \caption{Quantitative comparison of our method (FeudalNav and Stacked FeudalNav) against baselines and SOTA on the image goal task following the evaluation protocol from NRNS \cite{hahn2021no} in previously unseen Gibson environments \cite{xiazamirhe2018gibsonenv}. The top results are bolded for each category. We show a $69\%$ and $27\%$ increase in performance (success rate and SPL respectively) on curved trajectories on average over NRNS, and a $7\%$ increase in performance on the straight success rate. (* denotes the use of a simulator.)}
    \label{tab:NRNSTable}
    \vspace{-10pt}
\end{table*}

\begin{figure*}
    \centering
    \includegraphics[width=\linewidth]{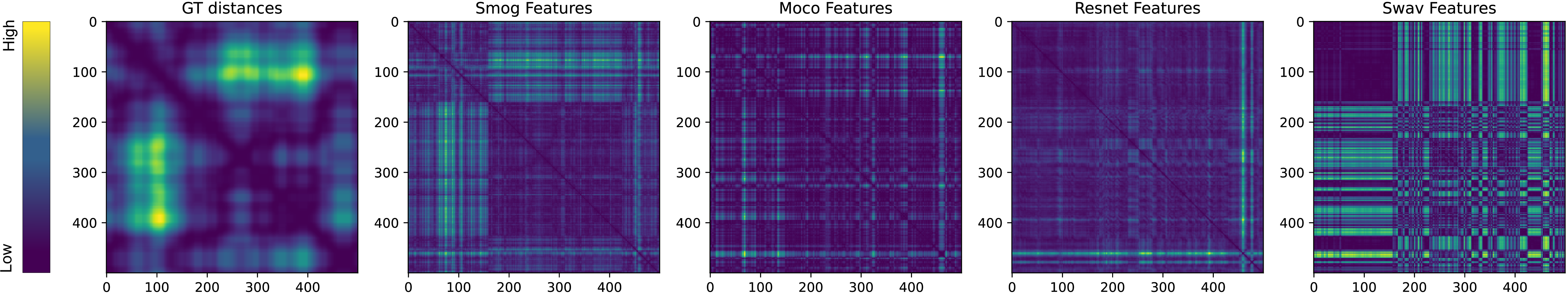}
    \vspace{-15pt}
    \caption{We compare feature distances between all pairs of images in an example trajectory from the human navigation dataset for different self-supervised contrastive learning methods against the ground truth distances for each pair of images. Lower feature distances are darker/purple and higher feature distances are brighter/yellow. We show that SMoG \cite{pang2022smog} learns a latent space that preserves ground truth distance trends the most, making it the most suitable for the \latentmap. }
    \vspace{-10pt}
    \label{fig:hlmFeats}
\end{figure*}

\section{Results}


\subsection{Mid-Level Manager Accurately Mimics Human Navigation}
Figure \ref{fig:mlmPoints} shows qualitative examples of the mid-level manager's predicted waypoints (blue) juxtaposed with the human ground truth point clicks (orange). For the majority of the points, both the predictions and the ground truth lie in the same area of the same object. When this is not true, the mid-level manager's predicted waypoints still lead to feasibly navigable areas such as other doorways or further into rooms, thus showing that they mimic human navigation.


\subsection{FeudalNav Outperforms Baselines and SOTA}

We show a comparison of our feudal navigation agent to SOTA in Table \ref{tab:NRNSTable}. We report success rate and SPL for FeudalNav, the version of our agent using the \midnet with a single RGBD-Map input, and Stacked FeudalNav, which uses the \midnet with three RGBD-Map inputs. Our method has shown a significant improvement in performance of $73\%$ (straight) and $158\%$ (curved) over all flat RL \cite{wijmans2019dd} baselines and  $157\%$ (straight) and $2341\%$ (curved) over both behavior cloning (BC) methods \cite{hahn2021no} while using no RL, learning no metric map, and not training directly in a simulator.

On average, we achieve a $7\%$ increase in performance on success rate over NRNS on the straight trajectories, despite not using odometry, not utilizing a graph, and only using $\sim37$k images for training (compared to the 3.5 million used by NRNS). Our SPL is consistently lower for straight trajectories because the hierarchical nature of our architecture prioritizes broader exploration \cite{nachum2019ask} more than other methodologies, which is a desirable trait when finding objects in previously unseen environments. This is evidenced by our consistent improvement of $69\%$ and $27\%$ percent over NRNS on both success rate \textbf{and} SPL respectively for curved trajectories of all difficulty levels. 
In the real world, it is less likely that a robot will be tasked to find an object within a straight line of sight from itself. For this reason, performance on the curved trajectory case gives a more realistic prediction of real world performance. 



\begin{table*}[]
    \centering
    \begin{tabular}{|c|c|c|c|c|c|c|c|c|c||c|c|}
    \hline
    \multirow{2}{5em}{Path Type} & \multirow{2}{3em}{\latentmapa} & \multirow{2}{7em}{\midneta} & \multirow{2}{3em}{Worker} & \multicolumn{2}{c|}{Easy} & \multicolumn{2}{c|}{Medium} & \multicolumn{2}{c||}{Hard} & \multicolumn{2}{c|}{Average}\\
     & & & & Succ$\uparrow$ & SPL$\uparrow$ & Succ$\uparrow$ & SPL$\uparrow$ & Succ$\uparrow$ & SPL$\uparrow$ & Succ$\uparrow$ & SPL$\uparrow$ \\
    \hline
    \hline
    \multirow{5}{4em}{Straight} & \xmark & RGB &  \cmark & 48.00 & 30.28 & 37.00 & 21.75 & 24.57 & 13.21 & 36.52 & 21.75 \\
    & \xmark & RGBD &  \cmark & 48.20 & 31.70 & 39.40 & 21.50 & 24.94 & 12.99 & 37.51 & 22.06 \\
    & \xmark & 3 RGBD &  \cmark & 50.20 & 31.19 & 38.60 & 21.24 & 20.72 & 9.16 & 36.51 & 20.53  \\
    & \cmark & RGBD-\latentmapa (FeudalNav) &  \cmark & 61.40 & 51.81 & 50.30 & 40.10 & 31.02 & 23.33 & 45.73 & 37.69 \\
    & \cmark & 3 RGBD-\latentmapa (Stacked FeudalNav) &  \cmark & 65.90 & 48.50 & 51.00 & 29.22 & 33.62 & 14.49 & 50.17 & 30.74 \\
    \hline
    \multirow{5}{4em}{Curved} & \xmark & RGB &  \cmark & 34.70 & 11.20 & 32.60 & 13.02 & 18.20 & 7.24 & 28.5 & 10.49  \\
    & \xmark & RGBD &  \cmark & 36.60 & 11.55 & 30.00 & 11.86 & 18.30 & 7.72 & 28.3 & 10.37  \\
    & \xmark & 3 RGBD &  \cmark & 39.50 & 11.91 & 32.80 & 11.75 & 15.70 & 5.65 & 29.33 & 9.77  \\
    & \cmark & RGBD-\latentmapa (FeudalNav)  &  \cmark & 41.30 & 19.51 & 32.60 & 17.10 & 18.60 & 10.88 & 30.83 & 15.83 \\
    & \cmark & 3 RGBD-\latentmapa (Stacked FeudalNav) &  \cmark & 56.40 & 21.37 & 44.10 & 17.41 & 21.00 & 7.30 & 40.50 & 15.36 \\
    \hline
    \end{tabular}
    \caption{We conduct an ablation study showing the effect of each module in FeudalNav on overall image-goal task performance. Despite similar success rate and SPL, we find a qualitative improvement as we add depth and 3 historical frames to the input to \midneta. We see large performance improvements as we add the \latentmap as input as it allows the network to have a notion of observation frequency and helps to localize the agent in the environment. }
    \label{tab:ablationTable}
    \vspace{-20pt}
\end{table*}

\section{Ablation Study}

\subsection{Different Network's Effects on the Memory Proxy Map}
We test the effectiveness of using multiple real world views as a contrastive learning augmentation.
The ground truth distances between pairs of observations across a full trajectory are compared to the distance between predicted image features for our high level manager network, which utilizes SMoG \cite{pang2022smog}. Moco \cite{he2020momentum}, Resnet-18 \cite{he2016deep}, and Swav \cite{caron2020unsupervised} are used as baselines for comparison. We compute inter-observation distance matrices in Figure \ref{fig:hlmFeats} showing the ground truth and predicted feature MSE distances for a single trajectory from the human navigation dataset from the Copemish Gibson environment. In the figure, lower distances are darker/purple and higher distances are lighter/yellow. The more a latent space preserves geometric distances between image features, the more it should resemble the first square showing the distance matrix created using ground truth distances. 

Swav \cite{caron2020unsupervised} learns features that are highly diverse, as evidenced by the large distances between the features and the abundance of yellow in the graph. Resnet \cite{he2016deep} has the reciprocal issue, where it learns highly similar features for all of the images in the trajectory, as evidenced by the large amount of purple in its graph. Both of these method's features are unsuitable for navigation applications because they do not provide useful environment information.  Moco \cite{he2020momentum} begins to mimic the GT distance matrix, but still mostly learns features that are similar to each other. SMoG \cite{pang2022smog} learns the most distance preserving latent space due to its ability to optimize inter-sample and inter-cluster distances simultaneously, making it the most useful for navigation purposes. 

\subsection{Full Hierarchy's Affect on Image-Goal Task Performance}

We also provide an ablation study to show how important each level of hierarchy is to our overall results in Table \ref{tab:ablationTable}. We incrementally add each piece of our architecture together and report the intermediary image-goal task results for the same experiment listed in Section V. The second column indicates whether or not the high level manager's \latentmap is included in the hierarchical navigation agent (\cmark) or not (\xmark). The third column indicates what type of input \midneta takes in to make its navigation waypoint predictions. We provide results for a single RGB image, a single RGBD image, three sequential RGBD images, the combined input of an RGBD image and the \latentmapa from the high level manager (FeudalNav), and the combined input of three RGBD images and three \latentmapa's (Stacked FeudalNav).

Despite their similar overall performance, we found that \midneta using RGB input performed qualitatively worse than using RGBD input because adding depth allowed the agent to avoid obstacles more efficiently. Again, we found that using three RGBD input images to \midneta also provided a qualitative performance boost to the agent as it stopped it from choosing to move in circles as it explored an environment. The added historical information imbued the agent with a notion of short term memory for recently visited locations that prompted it to explore new areas. 
We find a large improvement jump between using 3 RGBD images and a singular RGBD-Map input that is largely due to the introduction of memory into the navigation agent. The \latentmapa 1) gives the agent information about the frequency of its historical and current locations and 2) helps localize the agent with respect to the image features of its historical locations. We also show that using a 3 RGBD-Map input provides another large improvement in success rate corresponding to adding extra memory to the system. However, these same gains cannot be seen in the SPL which indicates that this extra historical context may be pushing the agent to over prioritize exploration.


\section{Conclusion and Discussion}

Many visual navigation methods, as well as the one presented here, are developed in simulation environments. The issue of transferring from simulation to real environments in visual navigation is an important one. Recent work \cite{gervet2023navigating} has empirically evaluated the sim2real performance of visual navigation methods and concluded that {\it modular} approaches succeeded well, but end-to-end frameworks did not transfer well to real environments. Our approach falls into this modular learning designation that facilitates real world transfer. 

With the ubiquity of advanced SLAM algorithms, the questions arise:  If SLAM is solved, why learn?, why use visual navigation?, what is there to learn in navigation?; why not just build a metric map?”. 
These are important questions to address. First, SLAM is often computationally intensive and requires a considerable amount of development time and compute resources to manage. Each new environment requires development effort to manage the resulting large-scale optimization problem. Visual navigation holds the promise of eventually being a light weight solution when no metric maps are built. But more importantly, visual navigation and learning-based navigation in general enables the system to learn about the dynamic environment that it is charged with navigating. The future challenges are learning 
nuances in navigating in social spaces such as: how close to other pedestrians, what speed, how to act in a crowd, and how to react to unexpected events.
These nuances are location-specific, culture-specific and are learned and refined readily by navigating humans, but providing the same capability to robots remains largely unsolved.

\section*{Acknowledgements}
This work was supported by the National Science Foundation (NSF) under grant NSF NRT-FW-HTF: Socially Cognizant Robotics for a Technology Enhanced Society (SOCRATES) No. 2021628 and grant nos. CNS-2055520, CNS-1901355, CNS-1901133.

\clearpage

\bibliographystyle{unsrtnat}
\bibliography{main}

\end{document}